\documentclass[10pt,twocolumn,letterpaper]{article}

\usepackage{iccv}
\usepackage{times}
\usepackage{epsfig}
\usepackage{graphicx}
\usepackage{amsmath}
\usepackage{amssymb}
\usepackage{dsfont}
\usepackage{color}
\usepackage{xcolor}
\usepackage{microtype} 
\usepackage{pifont}
\usepackage{amsmath}
\usepackage{amssymb}
\usepackage{dsfont}
\usepackage{bm}
\usepackage{xcolor,soul,colortbl}
\usepackage{array}
\usepackage{multirow}
\usepackage{subcaption}
\usepackage{dashrule}
\usepackage{graphicx}
\usepackage{breqn}
\usepackage{enumitem}

\setitemize{noitemsep,parsep=0pt,partopsep=0pt,topsep=0pt}
\setenumerate{noitemsep,parsep=0pt,partopsep=0pt,topsep=0pt}

\usepackage{xspace}
\usepackage{pifont}
\usepackage{booktabs}

\newcommand{\cmark}{\ding{51}}%
\newcommand{\xmark}{\ding{55}}%

\newcommand{\mcdropout}{MC Dropout\xspace}
\newcommand{\obsnet}{ObsNet\xspace}

\newcommand{\ab}[1]{\textcolor{black}{#1}}

\newcommand{\vic}[1]{#1}
\newcommand{\dap}[1]{#1}

\newcommand{\rem}[1]{}

\DeclareMathOperator{\sign}{sign}
\DeclareMathOperator*{\LAA}{LAA}

\newcommand{\parag}[1]{\smallskip\noindent\textbf{#1}~~}

\newcommand{\vx}{\boldsymbol{x}}

\usepackage[pagebackref=true,breaklinks=true,letterpaper=true,colorlinks,bookmarks=false]{hyperref}

\iccvfinalcopy 

\def\iccvPaperID{3734} 

\ificcvfinal\fi

\begin{document}

\title{Triggering Failures: Out-Of-Distribution detection by learning from local adversarial attacks in Semantic Segmentation}

\author{
Victor Besnier$^{1,3,4}$
\and 
Andrei Bursuc$^{2}$
\and 
David Picard$^{3}$ 
\and 
Alexandre Briot$^{1}$
\and 
1. Valeo, Créteil, France
\and 
2. Valeo.ai, Paris, France
\and 
3. LIGM, Ecole des Ponts, Univ Gustave Eiffel, CNRS, Marne-la-Vallée, France
\and
4. ETIS UMR8051, CY Université, ENSEA, CNRS, Cergy France

}

\maketitle
\ificcvfinal\thispagestyle{empty}\fi

\begin{abstract}
   \dap{In this paper, we tackle the detection of out-of-distribution (OOD) objects in semantic segmentation.
   By analyzing the literature, we found that current methods are either accurate or fast but not both which limits their usability in real world applications.
   To get the best of both aspects, we propose to mitigate the common shortcomings by following four design principles: decoupling the OOD detection from the segmentation task, observing the entire segmentation network instead of just its output, generating training data for the OOD detector by leveraging blind spots in the segmentation network and focusing the generated data on localized regions in the image to simulate OOD objects.
   Our main contribution is a new OOD detection architecture called ObsNet associated with a dedicated training scheme based on Local Adversarial Attacks (LAA). We validate the soundness of our approach across numerous ablation studies. We also show it obtains top performances both in speed and accuracy when compared to ten recent methods of the literature on three different datasets.}
\end{abstract}

\begin{figure}
\renewcommand{\captionfont}{\small}
\renewcommand{\captionlabelfont}{\bf}
    \centering 
    \includegraphics[clip, trim=5.2cm 16.5cm 7.5cm 4.0cm,width=0.9\linewidth]{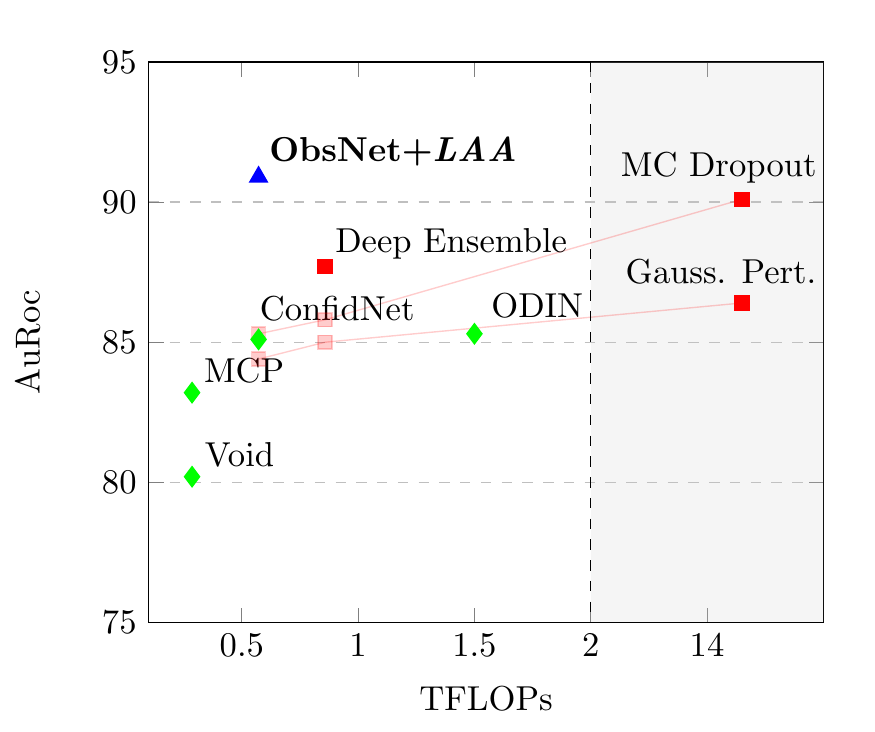}
    \vspace{-3mm}
    \caption{
    \ab{\textbf{Evaluation of precision vs. test-time computational cost on CamVid OOD.} 
    \ab{Existing methods for OOD detection in semantic segmentation are either accurate but slow (e.g., MC Dropout~\cite{Gal2016Dropout}, Deep Ensemble~\cite{Lakshminarayanan_2017}) or fast but inaccurate (e.g., Maximum Class Prediction~\cite{hendrycks17baseline}). In contrast, our method ObsNet+LAA is both accurate and fast.}
    } 
    \dap{Additional baselines and evaluation datasets are available in \S\ref{subsection:quantitative-results}.} 
    }
    \label{fig:teaser}
    \vspace{-4mm}
\end{figure}

\section{Introduction}

\ab{For real-world decision systems such as autonomous vehicles, accuracy is not the only performance requirement and it often comes second to reliability, robustness, and safety concerns~\cite{mcallister2017concrete}, as any failure carries serious consequences. Component modules of such systems frequently rely on Deep Neural Networks (DNNs) which have emerged as a dominating approach across numerous tasks and benchmarks \cite{szegedy2014going, he2016deep, he2017maskrcnn}. Yet, a major source of concern is related to the data-driven nature of DNNs as they do not always generalize to objects unseen in the training data. Simple uncertainty estimation techniques, e.g., entropy of softmax predictions~\cite{cover1999elements}, are less effective since modern DNNs are consistently overconfident on both in-domain~\cite{guo_2017} and out-of-distribution (OOD) data samples~\cite{nguyen2015deep,hendrycks17baseline,hein2019relu}. This hinders further the performance of downstream components relying on their predictions. Dealing successfully with the ``unknown unknown'', e.g., by launching an alert or failing gracefully, is crucial.}

\ab{In this work we address OOD detection for semantic segmentation, an essential and common task for visual perception in autonomous vehicles. \vic{We consider ``Out-of-distribution'', pixels from a region that has no training labels associated with. This encompasses unseen objects, but also noise or image alterations.} The most effective methods for OOD detection task stem from two major categories of approaches: ensembles and auxiliary error prediction modules. DeepEnsemble (DE)~\cite{Lakshminarayanan_2017} is a prominent and simple ensemble method that exposes potentially unreliable predictions by measuring the disagreement between individual DNNs. In spite of the outstanding performance, DE 
is computationally demanding for both training and testing and prohibitive for real-time on-vehicle usage.
For the latter category, given a trained main task network, a simple model is trained in a second stage to detect its errors or estimate its confidence~\cite{corbiere2019addressing,hecker2018failure,besnier2021learning}. Such approaches are computationally lighter, yet, in the context of DNNs, an unexpected drawback is related to the lack of sufficient negative samples, i.e., failures, to properly train the error detector~\cite{corbiere2019addressing}. This is due to an accumulation of causes: reduced size of the training set for this module (essentially a mini validation set to withhold a sufficient amount for training the main predictor), few mistakes made by the main DNNs, hence few negatives. 
}

\dap{In this work, we propose to} revisit the two-stage approach with modern deep learning tools in a semantic segmentation context. Given the application context, i.e., limited hardware and high performance requirements, we aim for reliable OOD detection (see \autoref{fig:teaser}) without compromising on predictive accuracy and computational time.
\ab{To that end we} \dap{introduce} \ab{four design principles aimed at mitigating the most common pitfalls and covering two main aspects, \textbf{(i)} \emph{architecture} and \textbf{(ii)} \emph{training}:}
    \\
\dap{    \indent \textbf{(i.a)} The pitfall of trading accuracy in the downstream segmentation task for robustness to OOD can be alleviated by decoupling OOD detection from segmentation.\\
    \indent \textbf{(i.b)} Since the processing performed by the segmentation network 
    \ab{aims to recognize known objects} and \ab{is} not adapted to OOD objects, the accuracy of the OOD detection can be improved significantly by observing the entire segmentation network instead of just its output.\\
    \indent  \textbf{(ii.a)} Training an OOD detector requires additional data that can be generated by leveraging blind spots in the segmentation network.\\
    \indent \textbf{(ii.b)} Generated data should focus on localized regions in the image to mimic unknown objects that are OOD.\\
}

\dap{Following these principles, we propose} a new OOD detection architecture called ObsNet and its associated training scheme based on Local Adversarial Attacks (LAA).
We experimentally show that \dap{our ObsNet+LAA method} achieves top performance in OOD detection on three semantic segmentation datasets (CamVid~\cite{BrostowFC:PRL2008}, StreetHazards~\cite{hendrycks_benchmark_2019} and BDD-Anomaly~\cite{hendrycks_benchmark_2019}), compared to a large set of methods\footnote{Code and data available at https://github.com/valeoai/obsnet}.

\parag{Contributions.} \dap{To summarize, our contributions are as follows:
    We propose a \textbf{new OOD detection method for semantic segmentation} based on four design principles: \ab{(i.a)} decoupling OOD detection from the segmentation task; \ab{(i.b)} observing the full segmentation network instead of just the output; \ab{(ii.a)} generating training data for the OOD detector using blind spots of the segmentation network; \ab{(ii.b)} focusing the adversarial attacks in localized region of the image to simulate unknown objects.
    We implement these four principles in a \textbf{new architecture called ObsNet} and its associated training scheme using \textbf{Local Adversarial Attacks (LAA)}.
    We perform \textbf{extensive ablation studies} on these principles to validate them empirically.
    We compare our method to \textbf{10 diverse methods} from the literature on \textbf{three datasets} (CamVid OOD, StreetHazards, BDD Anomaly) and we show it obtains \textbf{top performances both in accuracy and in speed}.
}

\parag{Strength and weakness.}\dap{The strengths and weaknesses of our approach are:
\begin{itemize}
    \item[\ding{51}] It can be used with any pre-trained segmentation network without altering their performances and without fine-tuning them \ab{(we train only the auxiliary module)}.
    \item[\ding{51}] It is fast since only 
    \ab{one extra} forward pass is required.
    \item[\ding{51}] It is very effective since we show it performs best compared to 10 very diverse methods from the literature on 
    \ab{three} different datasets.
    \item[\ding{55}] The pre-trained segmentation network has to allow for adversarial attacks, which is the case of commonly used deep neural networks.
    \item[\ding{55}] Our observer network has a memory/computation overhead equivalent to that of the segmentation network, which is not ideal for real time applications, but far less than that of MC Dropout or deep ensemble methods.
\end{itemize}
}

\dap{In the next section, we position our work with respect to the existing literature.}

\section{Related work}\label{section:related_work}
\begin{table*}
\renewcommand{\figurename}{Table}
\renewcommand{\captionlabelfont}{\bf}
\renewcommand{\captionfont}{\small} 
 \centering
  \begin{tabular}{ll@{\hspace{-0.25cm}}cccl} \toprule
    Type                &  Example                                   & OOD accuracy  & Fast Inference & Memory efficient     & Training specification           \\ \hline
    Softmax             &  MCP~\cite{hendrycks17baseline}            & -             & \checkmark     & \checkmark           & No                               \\
    Bayesian Learning   &  MC Dropout~\cite{Gal2016Dropout}          & \checkmark    & -              & \checkmark           & Reduces IoU acc.            \\ 
    Reconstruction      &  GAN~\cite{xia2020synthesize}              & \checkmark    & \checkmark     & \checkmark           & Unstable training \\
    Ensemble            &  DeepEnsemble~\cite{Lakshminarayanan_2017} & \checkmark    & -              & -                    & Costly Training             \\
    Auxiliary Network   &  ConfidNet~\cite{corbiere2019addressing}   & -             & \checkmark     & \checkmark           & Imbalanced train set     \\
    Test Time attacks   &  ODIN~\cite{Liang2018}                     & -*            & -              & \checkmark           & Extra OOD set               \\
    Prior Networks      &  Dirichlet~\cite{malinin2018}              & \checkmark    & \checkmark     & \checkmark           & Extra OOD set               \\ \hline
    \textbf{Observer} &  \textbf{ObsNet} + \textbf{LAA}              & \checkmark    & \checkmark     & \checkmark           & No                               \\ \bottomrule
  \end{tabular}
  \caption{\ab{\textbf{Summary of various OOD detection approaches amenable to semantic segmentation.} For real-time safety, key requirements for an OOD detector are accuracy, speed, easy training and memory efficiency.
  Our method addresses all requirements. Our \textit{LAA} is performed only at train time and mitigates the imbalance in the training data for the observer.}
  *Not accurate for semantic segmentation}
  \label{tab:recap_advantage}
  \vspace{-0.5em}
\end{table*}

\ab{The problem of data samples outside the original training distribution has been long studied for various applications before the deep learning era, under slightly different names and angles:}
outlier~\cite{breunig2000lof}, novelty~\cite{scholkopf2000support}, anomaly~\cite{liu2008isolation} and, more recently, OOD detection~\cite{hendrycks17baseline, hendrycks2018deep}. In the context of widespread DNN adoption this field has seen a fresh wave of approaches based on input reconstruction~\cite{schlegl2017unsupervised, baur2018deep, lis2019detecting, xia2020synthesize}, predictive uncertainty~\cite{Gal2016Dropout,KendallGal_2017, malinin2018}, ensembles~\cite{Lakshminarayanan_2017, franchi2019tradi}, adversarial attacks~\cite{Liang2018, lee2018simple}, using a \ab{void or} background class~\cite{ren2015faster, liu2016ssd} or dataset~\cite{bevandic2019simultaneous, hendrycks2018deep, malinin2018}, \etc., to name just a few. We outline here only some of the methods directly related to our approach \ab{and group them in a comparative summary in~\autoref{tab:recap_advantage}.}

\parag{Anomaly detection by reconstruction.}
In semantic segmentation, anomalies can be detected by training a (usually variational) autoencoder~\cite{creusot2015real,baur2018deep,venkataramanan2019attention} or generative model~\cite{schlegl2017unsupervised,lis2019detecting, xia2020synthesize} on in-distribution data. OOD samples are expected to lead to erroneous and less reliable reconstructions as they contain unseen patterns during training. On high resolution and complex urban images, autoencoders under-perform while more sophisticated generative models require large amounts of data to reach robust reconstruction \ab{or rich pipelines with re-synthesis and comparison modules.}

\parag{Bayesian approaches and ensembles.} 
BNNs~\cite{neal2012bayesian,Blundell2015} can capture predictive uncertainty 
\ab{from} distributions learned over 
network weights, \ab{but don't scale well~\cite{dusenberry2020efficient} 
and approximate solutions are preferred in practice.}
\ab{DE~}\cite{Lakshminarayanan_2017} is a highly effective, yet costly approach, that trains an ensemble of DNNs with different initialization seeds. 
\ab{Pseudo-ensemble approaches~\cite{Gal2016, maddox2019simple, franchi2019tradi, mehrtash2020pep} are a pragmatic alternative to DE that
bypass training of multiple networks and generate predictions from different random subsets of neurons~\cite{Gal2016, srivastava2014dropout} 
or from networks sampled from approximate weight distributions~\cite{maddox2019simple, franchi2019tradi, mehrtash2020pep}. However they all require multiple forward passes and/or storage of additional networks in memory.}
Our \obsnet is faster than ensembles as it requires only \dap{the equivalent of} two forward passes. Some approaches forego ensembling and propose deterministic networks that can output predictive distributions~\cite{malinin2018, sensoy2018, postels2019sampling, van2020uncertainty}. They typically trade predictive performance over computational efficiency and results can match \mcdropout~\cite{Gal2016Dropout} for uncertainty estimation.

\parag{OOD detection via test-time adversarial attacks.}
In ODIN, Liang \ab{et al.} \cite{Liang2018} leverage temperature scaling and small adversarial perturbations on the input at test-time to predict in- and out-of-distribution samples. 
\ab{Lee et. al~\cite{lee2018simple} extend this idea with a confidence score based on class-conditional Mahalanobis distance over hidden activation maps.}
\ab{Both approaches work best when train OOD data is available for tuning, yet this does not ensure generalization to other ODD datasets~\cite{shafaei2019less}. }
\dap{Contrarily to} us, ODIN uses adversarial attack \dap{at test time} as a \dap{\emph{method}} to detect OOD. However, so far this method has not been shown effective for structured output tasks where the test cost is likely to explode, as adversarial perturbations are necessary for each pixel. \dap{In contrast, we propose to use adversarial attacks during training as a \emph{proxy} for OOD training samples, with no additional test time cost.}

\parag{Learning to predict errors.}
Inspired by early approaches from model calibration literature~\cite{platt1999probabilistic, zadrozny2001obtaining, zadrozny2002transforming, naeini2015obtaining, naeini2016binary}, a number of methods propose endowing the task network with an error prediction branch allowing self-assessment of predictive performance. This branch can be trained jointly with the main network~\cite{devries2018learning, yoo2019learning}, however better learning stability and results are achieved 
\ab{with two-stage sequential training~\cite{corbiere2019addressing, hecker2018failure,besnier2021learning, Samson_2019_ICCV}}
Our \obsnet also uses an auxiliary network and is trained in two stages allowing it to learn from the failure modes of the task network. 
\ab{While}~\cite{corbiere2019addressing, hecker2018failure,besnier2021learning,Samson_2019_ICCV} 
focus on in-distribution errors, we address OOD detection for which there is no available training data.
\ab{In contrast with these methods that struggle with the lack of sufficient negative data to learn from, we devise an effective strategy to generate failures that further enable generalization to OOD detection.}
We redesign \ab{both} the training procedure and the architecture of the auxiliary network in order to deal with OOD examples, by introducing Local Adversarial Attack ($\LAA$). 

\parag{Generic approaches.} 
Finally we mention a set of mildly related approaches that do not address directly OOD detection, but achieve good performances on this task. In spite of the overconfidence pathological effect, using the maximum class probability from the softmax prediction can be used towards OOD detection~\cite{hendrycks17baseline, Oberdiek_2020_CVPR_Workshops}. Temperature scaling~\cite{guo_2017, platt1999probabilistic} is a strong post-hoc calibration strategy of the softmax predictions using a dedicated validation set. If predictions are calibrated, OOD samples can be detected by thresholding scores. Pre-training with adversarial attacked images~\cite{hendrycks2019using} has also been shown to lead to better calibrated predictions and good OOD detection for image classification. We consider these simple, yet effective approaches as baselines in order to validate the utility of our contribution. 

\section{Proposed Method}\label{section:method}


\dap{Following our analysis of the related work, we base our OOD semantic segmentation method on two categories of 
\ab{aspects:} \ab{\textbf{(i) Architecture:} OOD detection has to be decoupled from the 
segmentation prediction to retain maximal accuracy in both the segmentation and OOD task (\S\ref{subsection:ood_detector}); \textbf{(ii) Training:} Training an 
OOD detector without OOD data is difficult, but can be done nonetheless by generating training data with carefully designed adversarial attacks (\S\ref{subsection:laa}).}
}

\dap{Both of these 
\ab{aspects} require careful design to work effectively, which we detail in the following.
\ab{We validate them experimentally in \S\ref{section:results}.}
}


\begin{figure*}
\renewcommand{\captionfont}{\small}
\renewcommand{\captionlabelfont}{\bf}
    \centering
    \includegraphics[width=0.9\linewidth]{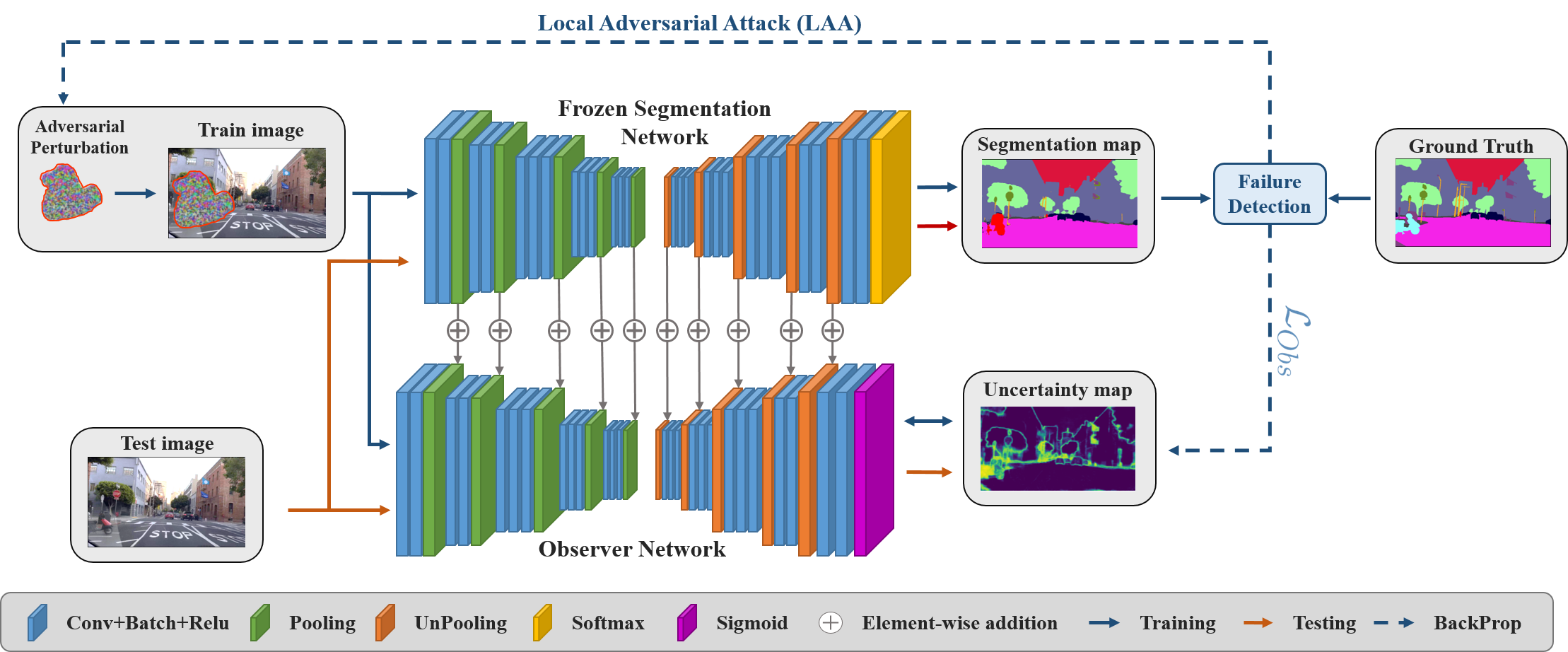}
    \caption{
    \ab{\textbf{Overview of our method.}} \textbf{Training (blue arrow)} The \emph{Segmentation Network} is frozen. The input image is perturbed by a local adversarial attack. Then the \emph{Observer Network} is trained to predict \emph{Segmentation Network}'s errors, given the images and some additional skip connections. \textbf{Testing (red arrow)} No augmentation is performed. The \emph{Observer Network} highlights the out-of-distribution sample, here a motor-cycle. To compute the uncertainty map, the \emph{Observer Network} requires only one additional forward pass compared to the standard segmentation prediction.}
    \label{fig:architecture}
    \vspace{-0.5em}
\end{figure*} 

\subsection{ObsNet: Dedicated OOD detector}
\label{subsection:ood_detector}
\dap{
Modifying the segmentation network to account for OOD is expected to impact its accuracy as we show in the experiments. Furthermore, it prevents from using off-the-shelf pretrained segmentation networks that have excellent segmentation accuracy.
As such, we follow a two-stage approach where an additional predictor tackles the OOD detection while the segmentation network remains untouched.
}

\dap{
In the literature, two-stage approaches are usually related to calibration \cite{platt1999probabilistic, zadrozny2001obtaining, zadrozny2002transforming, naeini2015obtaining, naeini2016binary} where the outputs of the segmentation network are 
\ab{mapped to} normalized scores. 
However this is not well adapted for segmentation since it does not use the spatial information contained in nearby predictions.
\ab{We} show in the experiments that using only the output of the segmentation network is not enough to obtain accurate OOD detection.
}

\dap{
As such, \ab{on the architecture side} we follow two design principles in our work: \\
\indent \ab{\textbf{(i.a)} OOD detection should be decoupled from the segmentation prediction to avoid any negative impact on the accuracy of the segmentation task.} \\
\indent \ab{\textbf{(i.b)} The OOD detector should observe the full segmentation network instead of just the output.}
}

\dap{We thus design an observer network called ObsNet that has a similar architecture to that of the segmentation network and attend the input, the output and intermediate feature maps of the segmentation network as shown on Figure \ref{fig:architecture}. We show experimentally that these design choices lead to increased OOD detection accuracy (see \S\ref{subsec:ablation}).
}


\dap{More formally, t}he observer network (denoted $Obs$) is trained to predict the probability that the segmentation network (\dap{denoted $Seg$)} output is not aligned with the correct class $y$:
\begin{equation}
  Obs(\vx, Seg_r(\vx))\approx Pr[ Seg(\vx) \neq y ],
\end{equation}
\ab{where $\vx$ is the input image and $Seg_r$ the} skip connections from intermediate feature maps of $Seg$. 


To that end, we train the ObsNet to minimize a binary cross-entropy loss function: 
\begin{multline}
\mathcal{L}_{Obs}(\vx, y) = (\mathds{1}_{Seg(\vx)\neq y}-1)\log (1 -Obs(\vx,Seg_r(\vx))) \\ 
 - \mathds{1}_{Seg(\vx) \neq y}\log Obs(\vx,Seg_r(\vx))
 \label{eq:obs_loss}
\end{multline}
\ab{with $\mathds{1}_{Seg(\vx) \neq y}$ the indicator function of $Seg(\vx) \neq y$.}

\dap{\parag{Discussion.}}
Since the observer network processes both the image and skip connections from the segmentation network, it has the ability to \emph{observe} internal behaviour and dynamics of $Seg$ which \ab{has} been shown to be different when processing an OOD samples (as measured by, \ab{e.g.,} Mahalanobis distance on feature maps~\cite{lee2018simple} or higher order Gram matrices on feature maps~\cite{sastry2019detecting}). 

We emphasize an advantage of our approach w.r.t. previous methods that is related to the low computational complexity, as we only have to make a single forward pass through the segmentation network and the observer network. Experimentally, \obsnet is 21 times faster than \mcdropout with 50 forward passes on a GeForce RTX 2080 Ti, 
while outperforming it 
\ab{(see \S\ref{section:results})}.
Moreover, our method can be readily used on state of the art pre-trained networks without requiring retraining or even fine-tuning them. 


\subsection{Training ObsNet by triggering Failures}
\label{subsection:laa}

\begin{figure*}
\renewcommand{\captionfont}{\small}
\renewcommand{\captionlabelfont}{\bf}
\renewcommand{\captionfont}{\small}
\renewcommand{\captionlabelfont}{\bf}
\centering
\begin{subfigure}{0.19\textwidth}
  \centering
  \includegraphics[height=6cm, width=\linewidth]{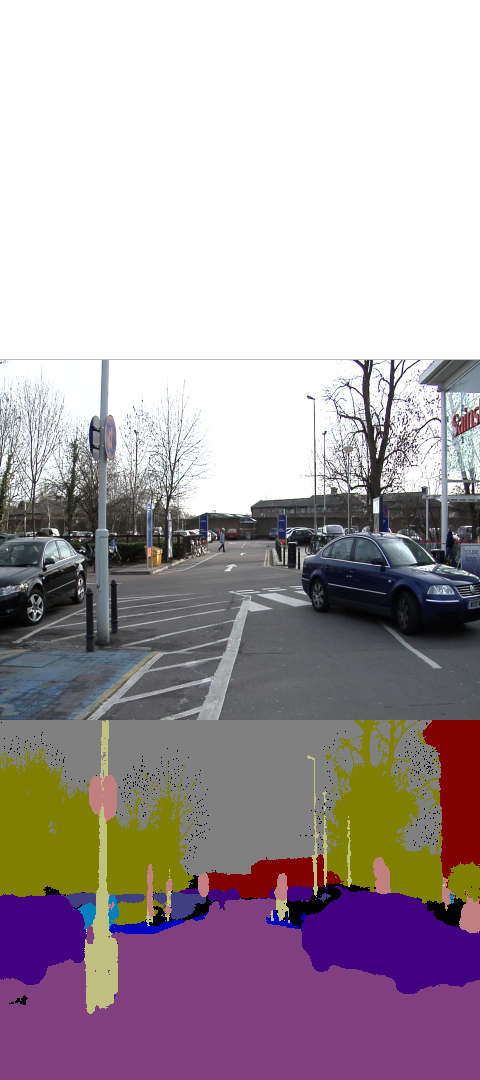}
  \caption{No attack}
\end{subfigure}
\begin{subfigure}{0.19\textwidth}
  \centering
  \includegraphics[height=6cm, width=\linewidth]{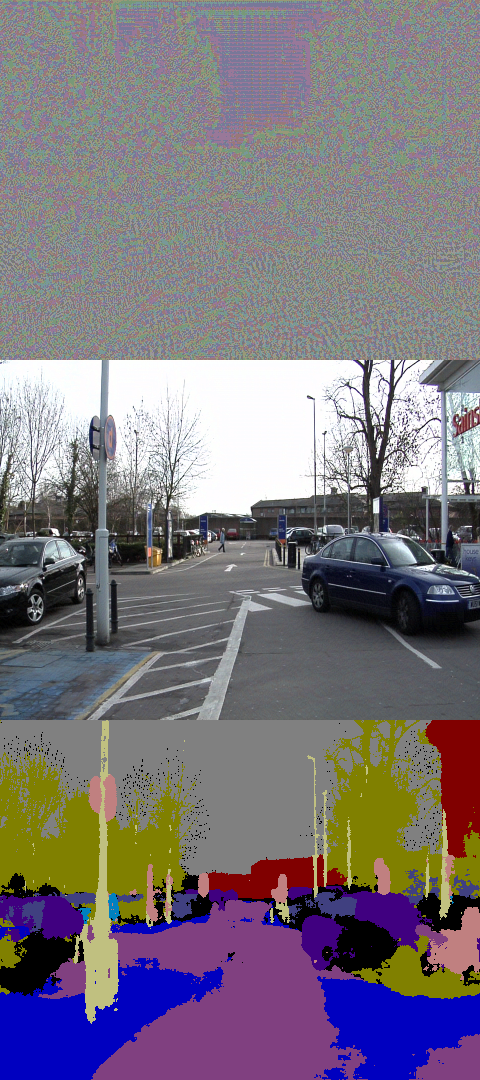}
  \caption{All pixels}
\end{subfigure}
\begin{subfigure}{0.19\textwidth}
  \centering
  \includegraphics[height=6cm, width=\linewidth]{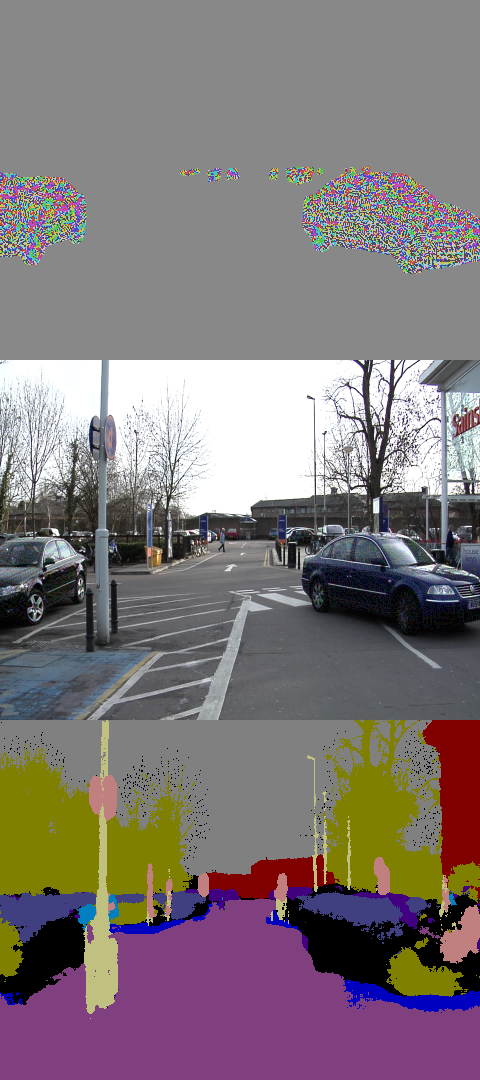}
  \caption{Class wise}
\end{subfigure}
\begin{subfigure}{0.19\textwidth}
  \centering
  \includegraphics[height=6cm, width=\linewidth]{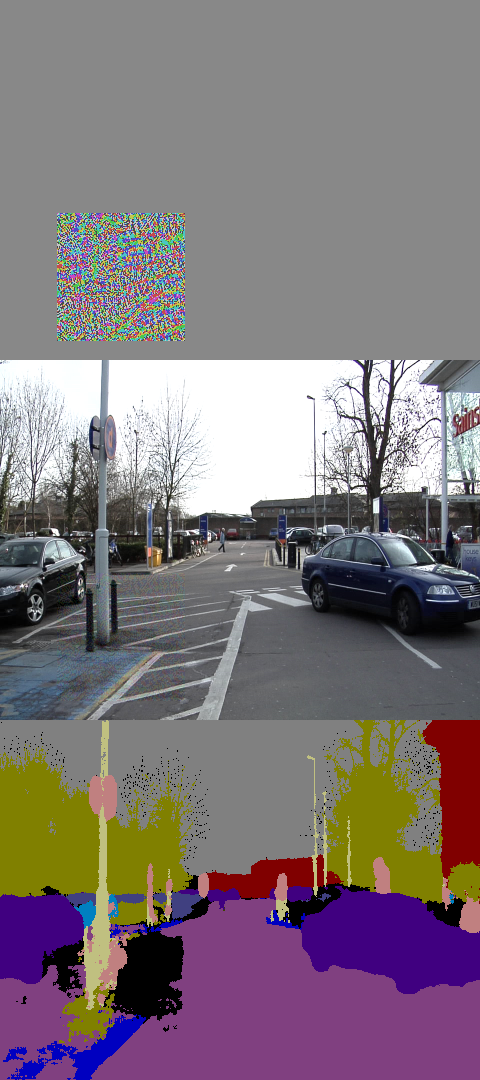}
 \caption{Square shape}
\end{subfigure}
\begin{subfigure}{0.19\textwidth}
  \centering
  \includegraphics[height=6cm, width=\linewidth]{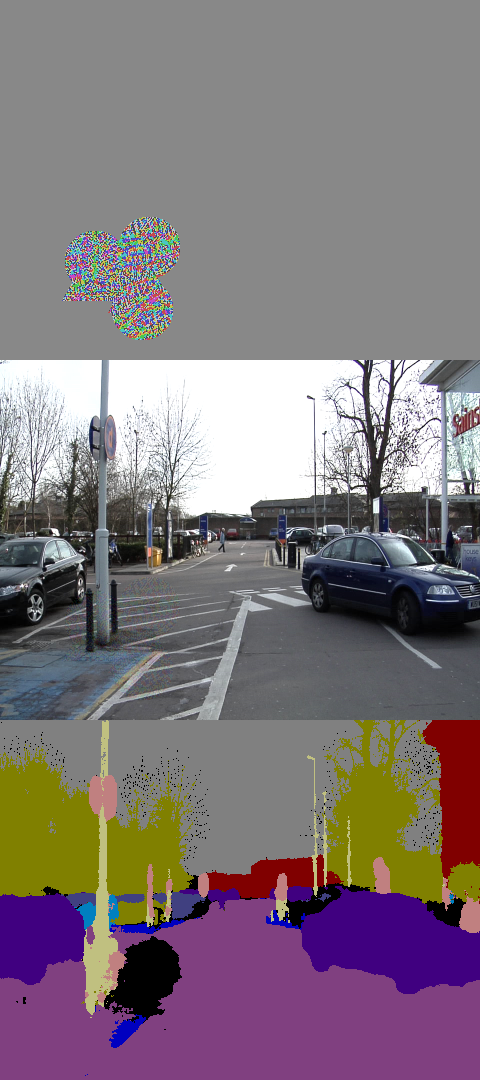}
  \caption{Random shape}
\end{subfigure}
\caption{
\ab{\textbf{Adversarial attack examples.}} 
\emph{Top}: Perturbations magnified 25$\times$; \emph{middle}: Input image with attacks; \emph{bottom}: SegNet prediction. 
}
\label{fig:ablation_noise}
\end{figure*}

\dap{
Without a dedicated training set of labeled OOD samples, one could argue that ObsNet is an error detector (similarly to \cite{corbiere2019addressing}) rather than an OOD detector and that it is furthermore very difficult to train since pre-trained segmentation networks are likely to make few errors.
We propose to solve both of these issues by following two design principles:\\
\indent \ab{\textbf{(ii.a)} The lack of training data should be tackled by generating training samples that trigger failures of the segmentation network, which we can obtain using adversarial attacks.}\\
\indent \ab{\textbf{(ii.b)} Adversarial attacks should be localized in space since OOD detection in a segmentation context corresponds to unknown objects.}  
}
\dap{We propose to generate the additional data required to train our ObsNet architecture by performing} Local Adversarial Attacks ($\LAA$) on the input image.
In practice, we select a region in the image by using a random shape and we perform a Fast Gradient Sign Method (FSGM)~\cite{Goodfellow2015} attack such that it is incorrectly classified by the segmentation network:
\begin{align}
    \Tilde{\vx} &= \vx + \LAA(Seg, \vx)\\
    \LAA(Seg, \vx) &= \epsilon \sign(\nabla_{\vx}\mathcal{L}(Seg(\vx),y))\Omega(\vx)
\end{align}
with step $\epsilon$, $\mathcal{L}(\cdot)$ the categorical cross entropy \dap{and $\Omega(\vx)$ the binary mask of the random shape}.
\ab{We show $\LAA$ examples in \autoref{fig:ablation_noise} and schematize the  training process in \autoref{fig:architecture}.}

The reasoning behind $\LAA$ is two-fold. First, by controlling the shape of the attack, we can make sure that the generated example does not accidentally belong to the distribution of the training set. 
Second, leveraging adversarial attacks allows us to focus the training just beyond the boundaries of the predicted classes which tend to be far from the training data due to the high capacity and overconfidence of DNNs, \dap{like OOD objects would be}. 

We show in the experiments that $\LAA$ produces a good training set for learning to detect OOD examples. In practice, we found that generating random shapes \dap{is essential to obtain good performances in contrast to non-local} adversarial attacks.
These random shapes coupled with $\LAA$ may mimic unknown objects or objects parts, exposing common behavior patterns in the segmentation network when facing them.
\ab{We validate our approach in an ablation study in \S\ref{subsec:ablation}}. 

\parag{Discussion.} 
We point out that by triggering failures using $\LAA$, we address the problem  of the low error rates of the segmentation network. We can in fact generate as many OOD\dap{-like} examples as needed to balance the positive (\ab{i.e.,} correct predictions) and negative (\ab{i.e.,} erroneous predictions) terms in \autoref{eq:obs_loss} for training the observer network. 
Thus, even if the segmentation network attains nearly perfect performances on the training set, we are still able to \dap{train the ObsNet to detect where the predictions of the segmentation network} are unreliable.

One could ask why not using $\LAA$ for training a more robust and reliable segmentation network in the first place, as done in previous works~\cite{Goodfellow2015, miyato2018virtual, hendrycks2019using}, instead of adding and training the observer network. Training with adversarial examples \dap{improves the robustness of the segmentation network at the cost of its accuracy (See \S\ref{subsec:ablation})}, but it will 
not make it infallible as there will still be numerous blind-spots in the multi-million dimensional parameter space of the network.
\dap{It also prevents from using pre-trained state-of-the-art segmentation networks.}
Here, we are rather interested in capturing the main failure modes of the 
segmentation network \ab{to enable \obsnet to learn and to recognize them later on OOD objects.}

\dap{Finally, one could ask why not perform adversarial attacks at test time as it is done in ODIN~\cite{Liang2018}. Performing test time attacks has two major drawbacks. First it is computationally intensive at test time since it requires numerous backward passes\ab{, i.e., one attack per pixel}. Second, it is not well adapted to segmentation as perturbations of a single pixels can have effect on a large areas (e.g., one pixel attacks) thus hindering the detection accuracy of perfectly valid predictions. We show in \S\ref{subsection:quantitative-results} that our training scheme is better performing both in accuracy and speed when compared to test time attacks.}

\section{Experiments}\label{section:results}
In this section, we present extensive experiments to validate that our proposed observer network combined with local adversarial attacks outperforms a large set of very different methods on three different benchmarks.

\subsection{Datasets \& Metrics}
To highlight our results, we select three datasets for Semantic Segmentation of urban streets scenes with anomalies in the test set. Anomalies correspond to out-of-distribution objects, not seen during train time.

\textbf{CamVid OOD}: We design a custom version of CamVid \cite{BrostowFC:PRL2008}, where we blit random animals from \cite{maboudi2008icpr} in a random part of the image. This dataset contains $367$ train and $233$ test images. There are $19$ different species of animals, and one animal in each test image. 
This setup is analog to that of Fishyscapes~\cite{blum2019fishyscapes}, with the main advantage that it does not require the use of an external evaluation server and that we provide a wide variety of baselines\footnote{\dap{To ensure easy reproduction and extension of our work}, we publicly release the code for dataset generation and model evaluation at https://github.com/valeoai/obsnet.}.

\textbf{StreetHazards}: This is a synthetic dataset \cite{hendrycks_benchmark_2019} from the Carla simulator. It is composed of $5125$ train and $1500$ test images, collected in six virtual towns. There are $250$ different kinds of anomalies (like UFO, dinosaur, helicopter, etc.) with at least one anomaly per image. 

\textbf{BDD Anomaly}: Composed of real images, this dataset is sourced from the BDD100K semantic segmentation dataset~\cite{bdd100k}. Here, motor-cycle and train are selected as anomalous objects and all images containing these objects are removed from the training set. The remaining dataset contains $6688$ images for training and $361$ for testing.

To evaluate each method on these datasets, we select three metrics for detecting misclassified and out-of-distribution examples and one metric for calibration:
\begin{itemize}
    \item[$\circ$] \textbf{fpr95tpr}~\cite{Liang2018}: It measures the false positive rate when the true positive rate is equal to 95\%. The aim is to obtain the lowest possible false positive rate while guaranteeing a given number of detected errors.
    \item[$\circ$] \textbf{Area Under the Receiver Operating Characteristic curve (AuRoc)}~\cite{hendrycks17baseline}: This threshold free metric corresponds to the probability that a certain example has a higher value than an uncertain one.
    \item[$\circ$] \textbf{Area under the Precision-Recall Curve (AuPR)}~\cite{hendrycks17baseline}: Also a threshold-independent metric. The AuPR is less sensitive to unbalanced dataset than AuRoc.
    \item[$\circ$] \textbf{Adaptive Calibration Error (ACE)}~\cite{Nixon_2019_CVPR_Workshops}: Compared to standard calibration metrics where bins are fixed, ACE adapts the range of each the bin to focus more on the region where most of the predictions are made.
\end{itemize}

For all our segmentation experiments we use a Bayesian SegNet \cite{badrinarayanan2015segnet}, \cite{kendall2015bayesian} as the main network. Therefore, our \obsnet follows the same architecture as this SegNet.
Ablation on the architecture of \obsnet, hyper-parameters and training details can be 
found in the supplementary material.

\subsection{Ablation Study}\label{subsec:ablation}
First, to validate that the local adversarial attack 
contributes to improving the observer network, we show on \autoref{eval_adv} the performance gap for each metric on each dataset. \dap{This validates the use of LAA to train the observer network as per principle (ii.a)}. 

\begin{table}
\renewcommand{\figurename}{Table}
\renewcommand{\captionlabelfont}{\bf}
\renewcommand{\captionfont}{\small} 
 \centering
  \begin{tabular}{l@{\hskip2.3mm}c@{\hskip2.3mm}c@{\hskip2.3mm}c@{\hskip2.3mm}c@{\hskip2.3mm}} \toprule
    Dataset      &  Adv          & fpr95tpr $\downarrow$ & AuPR $\uparrow$ & AuRoc $\uparrow$  \\ \hline
    CamVid OOD   &  \xmark       & 54.2                      & 97.1            & 89.1              \\
                 &  \cmark       & \textbf{44.6}             & \textbf{97.6}   & \textbf{90.9}     \\ \hline
    StreetHazards &  \xmark       & 50.1                      & 98.3            & 89.7              \\
                 &  \cmark       & \textbf{44.7}             & \textbf{98.9}   & \textbf{92.7}     \\ \hline
    BDD Anomaly  &  \xmark       & 62.4                      & 95.9            & 81.7              \\
                 &  \cmark       & \textbf{60.3}             & \textbf{96.2}   & \textbf{82.8}     \\ \bottomrule
  \end{tabular}
  \caption{Evaluation of the Local Adversarial Attack on each dataset.}
  \label{eval_adv}
\end{table}

The $\LAA$ can be seen as a data augmentation performed during \obsnet training. We emphasize that this type of data augmentation is not beneficial for the main network training, which is known as 
\ab{\emph{robust training}~\cite{MadryMSTV18}}, and that it requires an external observer network. Indeed, \autoref{tab_robust_training} illustrates the drop of accuracy when training the main network with the same adversarial augmentation as there is a trade-off between the accuracy and the robustness of a deep neural network \cite{TsiprasSETM19}. In contrast, our method keeps the main network frozen during \obsnet training, thus, the class prediction and the accuracy remain unchanged\dap{, validating principle (i.a)}.

\begin{table}
\renewcommand{\figurename}{Table}
\renewcommand{\captionlabelfont}{\bf}
\renewcommand{\captionfont}{\small} 
  \centering
  \begin{tabular}{lccc} \toprule
    Dataset       &  Robust         &  Mean IoU $\uparrow$  &  Global Acc $\uparrow$ \\ \hline
    Camvid ODD    &  -              &   \textbf{49.6}      &  \textbf{81.8}        \\
                  &  \checkmark     &   41.6      &  73.9        \\ \hline
    StreetHazards  &  -              &   \textbf{44.3}      &  \textbf{87.9}        \\ 
                  &  \checkmark     &   37.8      &  85.1        \\ \hline
    Bdd Anomaly   &  -              &   \textbf{42.9}      &  \textbf{87.0}        \\
                  &  \checkmark     &   41.5      &  85.9        \\ \bottomrule
  \end{tabular}
  \caption{Impact of robust training on accuracy.}
  \label{tab_robust_training}
\end{table}

In \autoref{tab_ablation_aa}, we show ablations on $\LAA$ by varying the type of noise (varying between attacking all pixels, random pixels, pixels from a specific class, pixels inside a square shape and pixels inside a random shape, see \autoref{fig:ablation_noise}). We conclude that local attacks on random shaped regions produce the best proxies for OOD detection (see supplementary material for detailed results)\dap{, validating principle (ii.b)}.

\begin{table}[t]
\renewcommand{\figurename}{Table}
\renewcommand{\captionlabelfont}{\bf}
\renewcommand{\captionfont}{\small} 
  \centering
  \begin{tabular}{lccc} \toprule
      Type                &  fpr95tpr $\downarrow$ &  AuPR $\uparrow$  &  AuRoc  $\uparrow$    \\   \hline
     All pixels           &  51.9                      &  97.1             &  89.6                       \\
     Sparse pixels        &  54.2                      &  97.2             &  89.6                  \\
     Class pixels         &  46.8                      &  97.2             &  89.9                         \\
     Square patch         &  45.5                      &  \textbf{97.4}    &  90.5                      \\
     Random shape         &  \textbf{44.6}             &  \textbf{97.4}    &  \textbf{90.6}             \\
    \bottomrule 
  \end{tabular}
  \caption{LAA ablation study by varying the attacked region.}
  \label{tab_ablation_aa}
\end{table}

In \autoref{tab_ablation_archia}, we conduct several ablation studies on the architecture of \obsnet. The main takeaway is that mimicking the architecture of the primary network and adding skip connections from several intermediate feature maps is essential to obtain the best performances (see full results in supplementary material)\dap{, validating principle (i.b)}.

\begin{table}[t!]
\renewcommand{\figurename}{Table}
\renewcommand{\captionlabelfont}{\bf}
\renewcommand{\captionfont}{\small} 
  \centering
  \begin{tabular}{l@{\hspace{-0.25cm}}cccc} \toprule
    Method               &  fpr95tpr $\downarrow$ &  AuPR $\uparrow$ &  AuRoc  $\uparrow$    \\ \hline
    Smaller architecture         &  60.3                      &  95.8           &  85.3                 \\ 
    \obsnet w/o skip    &  81.3                      &  92.0           &  74.4               \\
    \obsnet w/o input image               &  57.0                      &  96.9           &  88.2               \\
    \obsnet         &  \textbf{54.2}             &  \textbf{97.1}  &  \textbf{89.1}       \\ 
    \bottomrule                                                                                             
  \end{tabular}
  \caption{\obsnet architecture ablation study.}
  \label{tab_ablation_archia}
  \vspace{-0.5em}
\end{table}

\subsection{Quantitative and Qualitative results}
\label{subsection:quantitative-results}
We report results on \autoref{tab_camvid}, \autoref{tab_streethazard} and \autoref{tab_bddanomaly}, with all the metrics detailed above. We compare several methods:
\begin{itemize}
    \item[$\circ$] \textbf{MCP}~\cite{hendrycks17baseline}: Maximum Class Prediction. One minus the maximum of the prediction.
    \item[$\circ$] \textbf{AE}~\cite{hendrycks17baseline}: An autoencoder baseline. The reconstruction error is the uncertainty measurement.
    \item[$\circ$] \textbf{Void}~\cite{blum2019fishyscapes}: Void/background class prediction of the segmentation network.
    \item[$\circ$] \textbf{MCDA}~\cite{ayhan_test-time_2018}: Data augmentation such as geometric and color transformations is added during inference time. We use the entropy of 25 forward passes. 
    \item[$\circ$] \textbf{\mcdropout}~\cite{Gal2016Dropout}: The entropy of the mean softmax prediction with dropout. We use 50 forward passes for all the experiences.
    \item[$\circ$] \textbf{Gaussian Perturbation Ensemble}~\cite{franchi2019tradi,mehrtash2020pep}: We take a pre-trained network and perturb its weights with a random Normal distribution. This results in an ensemble of networks centered around the pre-trained model. 
    \item[$\circ$] \textbf{ConfidNet}~\cite{corbiere2019addressing}: ConfidNet is an observer network that is trained to predict the true class score. We use the code available online and modify the data loader to test ConfidNet on our experimental setup. 
    \item[$\circ$] \textbf{Temperature Scaling}~\cite{guo_2017}: We chose the hyper-parameters \emph{Temp} to have the best calibration on the validation set. Then, like MCP, we use one minus the maximum of the scaled prediction. 
    \item[$\circ$] \textbf{ODIN}~\cite{Liang2018}: ODIN performs test-time adversarial attacks on the primary network. We seek the hyper-parameters \emph{Temp} and $\epsilon$ to have the best performance on the validation set. The criterion is one minus the maximum prediction.
    \item[$\circ$] \textbf{Deep ensemble}~\cite{Lakshminarayanan_2017}: a small ensemble of 3 networks. We use the entropy the averaged forward passes.
\end{itemize}

As we can see on these tables, \obsnet significantly outperforms all other methods on detection metrics on all three datasets. Furthermore, ACE also shows that we succeed in having a good calibration value. 

\begin{table}
\renewcommand{\figurename}{Table}
\renewcommand{\captionlabelfont}{\bf}
\renewcommand{\captionfont}{\small} 
 \centering
  \begin{tabular}{l@{\hspace{-0.1cm}}c@{\hspace{0.1cm}}c@{\hspace{0.1cm}}c@{\hspace{0.1cm}}c@{\hspace{0.1cm}}} \toprule
    Method        & fpr95tpr$\downarrow$      & AuPR $\uparrow$   &  AuRoc $\uparrow$ & ACE $\downarrow$      \\ \hline        
    Softmax~\cite{hendrycks17baseline}                    &  65.4                      &  94.9             &  83.2             &  0.510              \\  
    Void~\cite{blum2019fishyscapes}                       &  66.6                      &  93.9             &  80.2             &  0.532              \\ 
    AE~\cite{hendrycks17baseline}                         &  93.0                      &  87.1             &  59.3             &  0.745              \\
    MCDA~\cite{ayhan_test-time_2018}                      &  66.5                      &  94.6             &  82.1             &  0.477              \\  
    Temp. Scale~\cite{guo_2017}                           &  63.8                      &  94.9             &  83.7             &  \textbf{0.356}     \\ 
    ODIN~\cite{Liang2018}                                 &  60.0                      &  95.4             &  85.3             &  0.500              \\ 
    ConfidNet~\cite{corbiere2019addressing}               &  60.9                      &  96.2             &  85.1             &  0.450              \\
    Gauss Pert.~\cite{franchi2019tradi,mehrtash2020pep}   &  59.2                      &  96.0             &  86.4             &  0.520              \\  
    Deep Ensemble~\cite{Lakshminarayanan_2017}            &  56.2                      &  96.6             &  87.7             &  0.459              \\ 
    \mcdropout~\cite{Gal2016Dropout}                      &  \underline{49.3}          &  \underline{97.3} &  \underline{90.1} &  0.463              \\    
    \textbf{\obsnet + \textit{LAA}}      &  \textbf{44.6}             &  \textbf{97.6}    &  \textbf{90.9}    &  \underline{0.446}  \\               
    \bottomrule                                                                                             
  \end{tabular}
  \caption{Evaluation on CamVid-ODD (best method in bold, second best underlined).}
  \label{tab_camvid}
  \vspace{-0.5em}
\end{table}

\begin{table}
\renewcommand{\figurename}{Table}
\renewcommand{\captionlabelfont}{\bf}
\renewcommand{\captionfont}{\small} 
  \centering
  \begin{tabular}{l@{\hspace{-0.1cm}}c@{\hspace{0.1cm}}c@{\hspace{0.1cm}}c@{\hspace{0.1cm}}c@{\hspace{0.1cm}}} \toprule
    Method              &  fpr95tpr $\downarrow$      &  AuPR $\uparrow$  &  AuRoc  $\uparrow$  &  ACE  $\downarrow$  \\ \hline
    Softmax~\cite{hendrycks17baseline}                     &  65.5                       &  94.7             &  80.8               &  0.463              \\ 
    Void~\cite{blum2019fishyscapes}                        &  69.3                       &  93.6             &  73.5               &  0.492              \\
    AE~\cite{hendrycks17baseline}                          &  84.6                       &  92.7             &  67.3               &  0.712              \\
    MCDA~\cite{ayhan_test-time_2018}                       &  69.9                       &  97.1             &  82.7               &  0.409              \\
    Temp. Scale~\cite{guo_2017}                            &  65.3                       &  94.9             &  81.6               &  \textbf{0.323}     \\
    ODIN~\cite{Liang2018}                                  &  61.3                       &  95.0             &  82.3               &  0.414              \\
    ConfidNet~\cite{corbiere2019addressing}                &  60.1                       &  98.1             &  90.3               &  0.399              \\
    Gauss Pert.~\cite{franchi2019tradi,mehrtash2020pep}    &  48.7                       &  98.5             &  90.7               &  0.449              \\
    Deep Ensemble~\cite{Lakshminarayanan_2017}             &  51.7                       &  98.3             &  88.9               &  0.437              \\
    \mcdropout~\cite{Gal2016Dropout}                       &  \underline{45.7}           &  \underline{98.8} &  \underline{92.2}   &  0.429              \\
    \textbf{\obsnet + \textit{LAA}}            &  \textbf{44.7}              &  \textbf{98.9}    &  \textbf{92.7}      &  \underline{0.383}  \\ \bottomrule
    \end{tabular}
  \caption{Evaluation on StreetHazard (best method in bold, second best underlined).}
  \label{tab_streethazard}
  \vspace{-0.5em}
\end{table}

\begin{table}
\renewcommand{\figurename}{Table}
\renewcommand{\captionlabelfont}{\bf}
\renewcommand{\captionfont}{\small} 
  \centering
  \begin{tabular}{l@{\hspace{-0.1cm}}c@{\hspace{0.1cm}}c@{\hspace{0.1cm}}c@{\hspace{0.1cm}}c@{\hspace{0.1cm}}} \toprule
    Method              &  fpr95tpr $\downarrow$      &  AuPR $\uparrow$  & AuRoc  $\uparrow$   &  ACE  $\downarrow$  \\ \hline
    Softmax~\cite{hendrycks17baseline}                     &  63.5                       &  95.4             & 80.1                &  0.633              \\
    Void~\cite{blum2019fishyscapes}                        &  68.1                       &  92.4             & 75.3                &  0.499              \\
    AE~\cite{hendrycks17baseline}                          &  92.1                       &  88.0             & 53.1                &  0.832              \\
    MCDA~\cite{ayhan_test-time_2018}                       &  61.9                       &  95.8             & 82.0                &  0.411              \\
    Temp. Scale~\cite{guo_2017}                            &  61.8                       &  95.8             & 81.9                &  \textbf{0.287}     \\
    ODIN~\cite{Liang2018}                                  &  \underline{60.6}           &  95.7             & 81.7                &  0.353              \\
    ConfidNet~\cite{corbiere2019addressing}                &  61.6                       &  95.9             & 81.9                &  0.367              \\
    Gauss Pert.~\cite{franchi2019tradi,mehrtash2020pep}    &  61.3                       &  96.0             & 82.5                &  0.384              \\
    Deep Ensemble~\cite{Lakshminarayanan_2017}             &  \textbf{60.3}              &  \underline{96.1} & 82.3                &  0.375              \\
    \mcdropout~\cite{Gal2016Dropout}                       &  61.1                       &  96.0             & \underline{82.6}    &  0.394              \\
    \textbf{\obsnet + \textit{LAA}} &  \textbf{60.3}              &  \textbf{96.2}    & \textbf{82.8}       &  \underline{0.345}  \\
    \bottomrule                                                                                             
  \end{tabular}
  \caption{Evaluation on Bdd Anomaly (best method in bold, second best underlined).}
  \label{tab_bddanomaly}
  \vspace{-1em}
\end{table}


\begin{figure*}
\renewcommand{\captionfont}{\small}
\renewcommand{\captionlabelfont}{\bf}
\centering
\includegraphics[width=\linewidth]{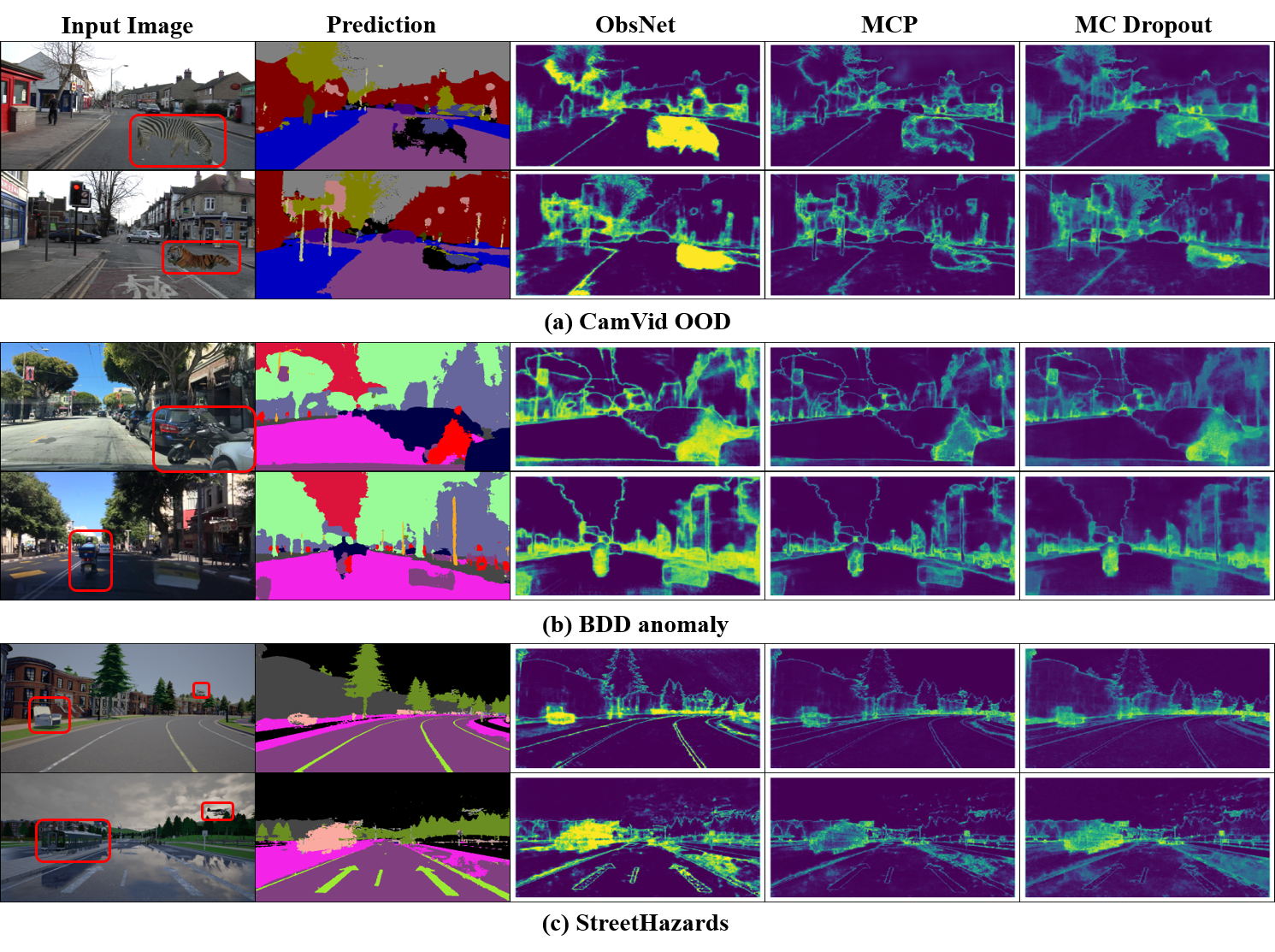}
\caption{
\vic{Uncertainty map visualization. \textbf{1st column:} We highlight the ground truth locations of the OOD objects to help visualize them (red bounding box). \textbf{2nd column:} Segmentation map of the SegNet. \textbf{3rd to 5th columns:} Uncertainty Map highlight in yellow. Our method produces stronger responses on OOD regions compared to other methods, while being as strong on regular error regions, e.g., boundaries}}
\vspace{-0.5em}
\label{fig:image_prez}
\end{figure*}

To show where the uncertainty is localized, we outline the uncertainty map on the test set (see \autoref{fig:image_prez}). 
We can see that our method is not only able to correctly detect OOD objects, but also to highlight areas where the predictions are wrong (edges, small and far objects, etc).


\dap{Finally, the trade-off between accuracy and speed is shown on \autoref{fig:teaser}, where we obtain excellent accuracy without any compromise over speed.}
\section{Conclusion}
In this paper, we propose an observer network called \obsnet to address OOD detection in semantic segmentation, by learning from triggered failures. We use skip connection to allow the observer network to seek abnormal behaviour inside the main network. We use local adversarial attacks to trigger failures in the segmentation network and train the observer network on these samples. We show on three different segmentation datasets that our strategy combining an observer network with local adversarial attacks is fast, accurate and is able to detect unknown objects.

\clearpage
{\small
\bibliographystyle{ieee_fullname}
\bibliography{section/citation}
}

\clearpage
\setcounter{section}{0}
\section{\ab{Implementation details \& hyper-parameters}}
For our implementation, we use Pytorch\footnote{A Paszke et al., \emph{PyTorch: An Imperative Style, High-Performance Deep Learning Library}, NIPS 2019} and will release the code after the review. We share each hyper-parameter in \autoref{tab:Hyperparameters}. We train \obsnet 
with SGD with momentum and weight decay 
\ab{for at most $50$ epochs using early-stopping.}
\obsnet is not 
\ab{trained} from scratch as we initialize the weights with those of the segmentation network. We also use a scheduler to divide the learning rate by 2 at epoch 25 and epoch 45. We use the same data augmentation (\ie Horizontal Flip and Random Crop) 
\ab{for} training of the segmentation network and 
\ab{as well as for} \obsnet.
As there are few errors in the training of \obsnet, we increase the weight of positive examples in the loss contribution (Pos Weight in \autoref{tab:Hyperparameters}).

\begin{table}[h!]
    \centering
    \begin{tabular}{l@{\hskip2.3mm}l@{\hskip2.3mm}l@{\hskip2.3mm}l@{\hskip2.3mm}} \toprule
    Params               & CamVid               & StreetHazards         & Bdd Anomaly            \\ \hline
    Epoch                & 50                   & 50                    & 50                     \\     
    Optimizer            & SGD                  & SGD                   & SGD                    \\ 
    LR                   & 0.05                 & 0.02                  & 0.02                   \\
    Batch Size           & 8                    & 6                     & 6                      \\
    Loss                 & BCE                  & BCE                   & BCE                    \\
    Pos Weight           & 2                    & 3                     & 3                      \\
    LAA shape            & rand shape           & rand shape            & rand shape             \\
    LAA type             & \ab{$\min_{p(c)}$}         & \ab{$\max_{p(k \neq c)}$}   & \ab{$\max_{p(k \neq c)}$}    \\
    epsilon              & 0.02                 & 0.001                 & 0.001                  \\ \bottomrule
    \end{tabular}
    \caption{
    \ab{Hyper-parameters to train} \obsnet on the different datasets.}
    \label{tab:Hyperparameters}
\end{table}

\begin{table}[t!]
  \centering
  \begin{tabular}{l@{\hspace{-0.09cm}}l@{\hskip0.01mm}c@{\hspace{0.1cm}}c@{\hspace{0.1cm}}c@{\hspace{0.1cm}}c@{\hspace{0.1cm}}} \toprule
                         & Type                &  fpr95tpr $\downarrow$ &  AuPR $\uparrow$  &  AuRoc  $\uparrow$   &  ACE  $\downarrow$     \\   \hline
                         & \mcdropout          &  49.3                      &  97.3             &  90.1                &  0.463               \\
                         & \obsnet base     &  54.2                      & 97.1              &  89.1                &  0.396               \\   \hline
                     
                         & all pixels           &  53.2                      &  97.1             &  89.5                &  0.410               \\
                         & sparse pixels       &  61.1                      &  97.1             &  89.2                &  \textbf{0.387}      \\
    \hspace{-0.12cm}\ab{$\min_{p(c)}$}          & class pixels              &  45.6                      &  97.3             &  90.3                &  0.428           \\
     & square patch        &  47.4                      &  97.3             &  90.1                &  0.461               \\
                         & rand shape        &  \textbf{44.6}             &  \textbf{97.6}    &  \textbf{90.9}       &  0.446               \\   \hline
    
                         & all pixels           &  51.9                      &  97.1             &  89.6                &  0.405               \\
                         & sparse pixels        &  54.2                      &  97.2             &  89.6                &  \textbf{0.374}      \\
    \hspace{-0.12cm}\ab{$\max_{p(k \neq c)}$}  & class pixels               &  46.8                      &  97.2             &  89.9                &  0.432           \\
                     & square patch        &  45.5                      &  \textbf{97.4}    &  90.5                &  0.464               \\
                         & rand shape        &  \textbf{44.6}             &  \textbf{97.4}    &  \textbf{90.6}       &  0.446               \\
    \bottomrule                                                                                             
  \end{tabular}
  \caption{Ablation \ab{on} adversarial attacks.}
  \label{tab_ablation_full}
\end{table}

\section{Ablation on \obsnet architecture, $\epsilon$ and $\LAA$}
One contribution of our work is the ablation we do on the architecture of the observer network compared to previous methods. We highlight that the skip connections are \ab{essential for reaching best performance}. For the smaller architecture, instead of keeping the same architecture as the segmentation network, we design \ab{a smaller variant:} a
\ab{convolutional} network with three 
\ab{convolutional} layers and a fully connected 
\ab{layer}. This architecture 
\ab{mimicks} the one used by ConfidNet [10].  
\begin{table}[t!]
  \centering
  \begin{tabular}{lc@{\hspace{0.15cm}}c@{\hspace{0.15cm}}c@{\hspace{0.15cm}}c@{\hspace{0.15cm}}} \toprule
    Method          &  fpr95tpr $\downarrow$ &  AuPR $\uparrow$ &  AuRoc  $\uparrow$   &  ACE  $\downarrow$    \\ \hline
    Smaller archi.     &  60.3                      &  95.8           &  85.3                &  0.476                 \\ 
    w/o skip     &  81.3                      &  92.0           &  74.4                &  0.551                 \\
    w/o input img  &  57.0                      &  96.9           &  88.2                &  0.455                 \\
    w/o pretrain   &  55.7                      &  96.9           &  88.7                &  0.419                 \\
    \obsnet full             &  \textbf{54.2}             &  \textbf{97.1}  &  \textbf{89.1}       &  \textbf{0.396}        \\ 
    \bottomrule                                                                                             
  \end{tabular}
  \caption{Ablation \obsnet without $\LAA$ training.}
  \label{tab_ablation_archib}
\end{table}

Next, we outline most of the experiments we make on $\LAA$. First, there are two different kinds of setups, we can either minimize the prediction class (\ie \ab{$\min_{p(c)}$}) or maximize \ab{instead} a different class (\ie \ab{$\max_{p(k \neq c)}$}), with $p=Seg(x)$ the class vector, \ab{$c = \max_{p}$} the maximum class prediction and $k$ a random class. Then, we attack with five different strategies: all pixels in the image, random sparse pixels, the area of a random class, all pixels in a square patch and all pixels in a random shape. We show \ab{in} \autoref{tab_ablation_full} the complete result\ab{s} on CamVid ODD. We can see that \ab{that random shape is the most effective.}  We use the FSGM because it’s a well-known and easy-to-use adversarial attack. Since our goal is to hallucinate OOD objects, we believe the location and the shape of the attacked region are the important part.

\begin{table}[t!]
\renewcommand{\figurename}{Table}
\renewcommand{\captionlabelfont}{\bf}
\renewcommand{\captionfont}{\small} 
 \centering
  \begin{tabular}{l@{\hspace{-0.1cm}}c@{\hspace{0.1cm}}c@{\hspace{0.1cm}}c@{\hspace{0.1cm}}c@{\hspace{0.1cm}}} \toprule
    Method        & fpr95tpr $\downarrow$      & AuPR $\uparrow$   &  AuRoc $\uparrow$ & ACE $\downarrow$       \\ \hline
    Softmax [25]      & 61.9                       & 96.5              & 84.4              &  0.480                 \\  
    Void [6]          & 79.9                       & 90.7              & 67.3              &  0.504                 \\  
    MCDA [1]          & 65.8                       & 96.3              & 83.1              &  0.440                 \\  
    Temp. Scale [19]  & 61.9                       & 96.6              & 84.6              &  \textbf{0.302}        \\ 
    ODIN [32]          & 58.3                       & 97.2              & 87.9              &  0.478                 \\ 
    ConfidNet [10]    & \underline{52.2}           & 97.5              & \underline{88.6}  &  0.412                 \\
    Gauss Pert. [15,41]  & 60.2                       & 96.8              & 85.6              &  0.497                 \\  
    Deep Ensemble [30] & 55.3                       & 97.5              & 88.1              &  \underline{0.343}     \\ 
    \mcdropout [17]   & 52.5                       & \underline{97.9}  & 88.5              &  0.443                 \\
    \obsnet + \textit{LAA}      & \textbf{47.7}              & \textbf{98.1}     & \textbf{90.3}     &  0.370                 \\ \bottomrule                                                                                             
  \end{tabular}
  \caption{Error detection evaluation on CamVid (best method in bold, second best underlined).}
  \label{tab_camvid_base}
\end{table}

\begin{figure}[t!]
\renewcommand{\figurename}{Figure}
\renewcommand{\captionlabelfont}{\bf}
\renewcommand{\captionfont}{\small} 
    \centering 
    \includegraphics[clip, trim=5.2cm 9cm 7.5cm 11.4cm,width=\linewidth]{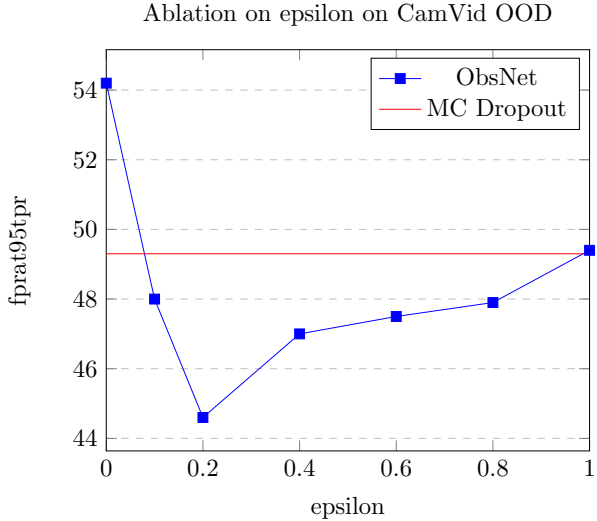}
    \caption{Evolution of the Fpr at 95 Tpr for different values of epsilon on CamVid OOD.}
    \label{fig:epsilon}
\end{figure}

As shown on \autoref{fig:epsilon}, we can see that the best $\epsilon$ for the attack is $0.02$ with a random shape blit at a random \ab{position in the image.} We can also see that even with a large $\epsilon$, \obsnet \ab{still achieves reasonable} performance.

\section{Error detector}
The observer is trained to assess whether the prediction differs from the true class (which is always the case for OOD regions), so it also tends to assign low conﬁdence scores for in-domain regions with likely high errors, as shown in \autoref{fig:error_detection}. This behavior is not caused by ObsNet, but depends on the accuracy of the main network at test time and should lessen with more accurate networks.
This effect shows that our method can be used for error detection, and outperforms all other methods, as illustrated in \autoref{tab_camvid_base}.

\section{\vic{Additional Experiments: DeepLab v3+}}
\vic{
We show on \autoref{tab:deeplab}, the results on BDD Anomaly with a more recent Deeplab v3+\footnote{LC Chen et al., \emph{Encoder-Decoder with Atrous Separable Convolution for Semantic Image Segmentation}, ECCV 2018} with ResNet-101 encoder. Our methods performs the best, while methods like ConfidNet do not scale when the segmentation accuracy increases as they have fewer errors to learn from.
}  
\begin{table}[h!]
\renewcommand{\figurename}{Table}
\renewcommand{\captionlabelfont}{\bf}
\renewcommand{\captionfont}{\small} 
  \centering
 \vspace{-2mm}
  \begin{tabular}
  {l@{\hspace{-0.1cm}}c@{\hspace{0.1cm}}c@{\hspace{0.1cm}}c@{\hspace{0.1cm}}c@{\hspace{0.1cm}}} \toprule
    Method                            &  fpr95tpr $\downarrow$  &  AuPR $\uparrow$      & AuRoc  $\uparrow$   &  ACE  $\downarrow$   \\ \hline
    Softmax [25]                     &  60.3                   &  95.8                 & 81.4                &  0.228               \\
    Void [6]                          &  68.8                   &  90.2                 & 74.0                &  0.485               \\
    MCDA [1]                        &  68.1                   &  95.1                 & 78.8                &  0.265               \\
    ConfidNet [10]                     &  64.5                   &  95.4                 & 80.9                &  0.254               \\
    Gauss Pert. [15,41]              &  61.4                   &  \underline{96.1}     & \underline{82.4}    &  \underline{0.186}   \\
    \mcdropout [17]                   &  \underline{60.0}       &  96.0                 & 82.0                &  0.219               \\
    \textbf{\obsnet + \textit{LAA}}   &  \textbf{58.8}          &  \textbf{96.3}        & \textbf{83.0}       &  \textbf{0.185}      \\
    \bottomrule   
    \end{tabular}
  \caption{Evaluation on Bdd Anomaly (best method in bold, second best underlined), with a DeepLab v3+.}
  \label{tab:deeplab}
  \vspace{-1em}
\end{table}

\section{CamVid OOD dataset}
For our experiments, we use urban street segmentation datasets 
\ab{with} anomalies \ab{withheld during training}. Unfortunately, there are few datasets with anomalies in the test set. 
\ab{For this reason} we propose the \ab{CamVid OOD} that will be made public after the review. To design \ab{CamVid OOD}, we blit random animals 
\ab{in test images of CamVid}. 
\ab{We add one different such anomaly in each of the $233$ test images.}
The rest of \ab{ the $367$} training images remain unchanged. The \ab{anomalous} animals are
\ab{\emph{bear}, \emph{cow}, \emph{lion}, \emph{panda}, \emph{deer}, \emph{coyote}, \emph{zebra}, \emph{skunk}, \emph{gorilla}, \emph{giraffe}, \emph{elephant}, \emph{goat}, \emph{leopard}, \emph{horse}, \emph{cougar}, \emph{tiger}, \emph{sheep}, \emph{penguin}, and \emph{kangaroo}.}
Then, we add \ab{them to }a $13$th class which is 
\ab{\emph{animals}/\emph{anomalies}} 
as the corresponding ground truth of the test set.

\begin{table}
\renewcommand{\figurename}{Table}
\renewcommand{\captionlabelfont}{\bf}
\renewcommand{\captionfont}{\small} 
 \centering
 \vspace{-2.5mm}
  \begin{tabular}{l@{\hspace{-0.1cm}}c@{\hspace{0.1cm}}c@{\hspace{0.1cm}}c@{\hspace{0.1cm}}c@{\hspace{0.1cm}}} \toprule
    Method        & fpr95tpr $\downarrow$      & AuPR $\uparrow$   &  AuRoc $\uparrow$ & ACE $\downarrow$ \\ \hline
    Softmax [25]      & 67.5                       & 94.7              & 82.5              & 0.529            \\  
    ConfidNet [10]    & 58.4                       & 96.4              & 86.8              & 0.462            \\
    Gauss Pert. [15,41]   & 61.8                       & 95.8              & 85.7              & 0.473            \\  
    Deep Ensemble [30] & 63.9                       & 96.5              & 86.4              & 0.468            \\ 
    \mcdropout [17]   & 52.8                       & 97.2              & 88.5              & 0.483            \\
    \obsnet + \textit{LAA}      & \textbf{42.1}              & \textbf{97.7}     & \textbf{91.4}     & \textbf{0.423}   \\ \bottomrule                                                                                             
  \end{tabular}
  \caption{Error detection evaluation on CamVid with random square attacks (best method in bold).}
  \label{tab_eval_adv}
  \vspace{-1em}
\end{table}

\begin{figure*}
\renewcommand{\figurename}{Figure}
\renewcommand{\captionlabelfont}{\bf}
\renewcommand{\captionfont}{\small} 
    \centering
    \includegraphics[width=\linewidth]{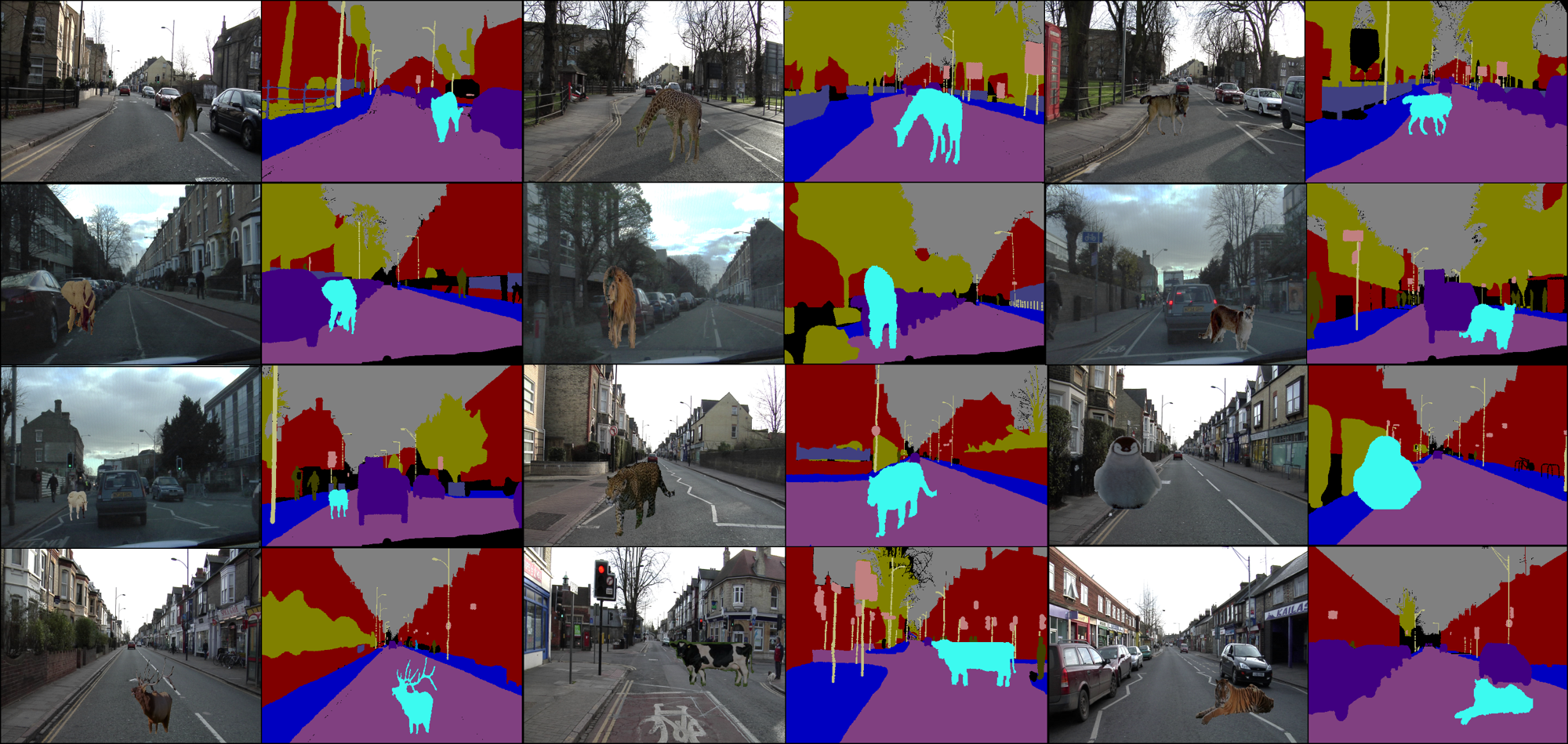}
    \caption{Examples of our dataset with anomalies and the ground truth.}
    \label{fig:camvid_odd_ex}
    \vspace{-0.5em}
\end{figure*}

This setup is similar to the Fishyscape dataset [6], without the constraint of sending a Tensorflow model online for evaluation. Thus, our dataset is easier to work with. We present some examples of the anomalies in \autoref{fig:camvid_odd_ex} with the ground truth highlighted in cyan.

\begin{figure*}[t!]
\renewcommand{\figurename}{Figure}
\renewcommand{\captionlabelfont}{\bf}
\renewcommand{\captionfont}{\small} 
    \centering
    \includegraphics[width=\linewidth]{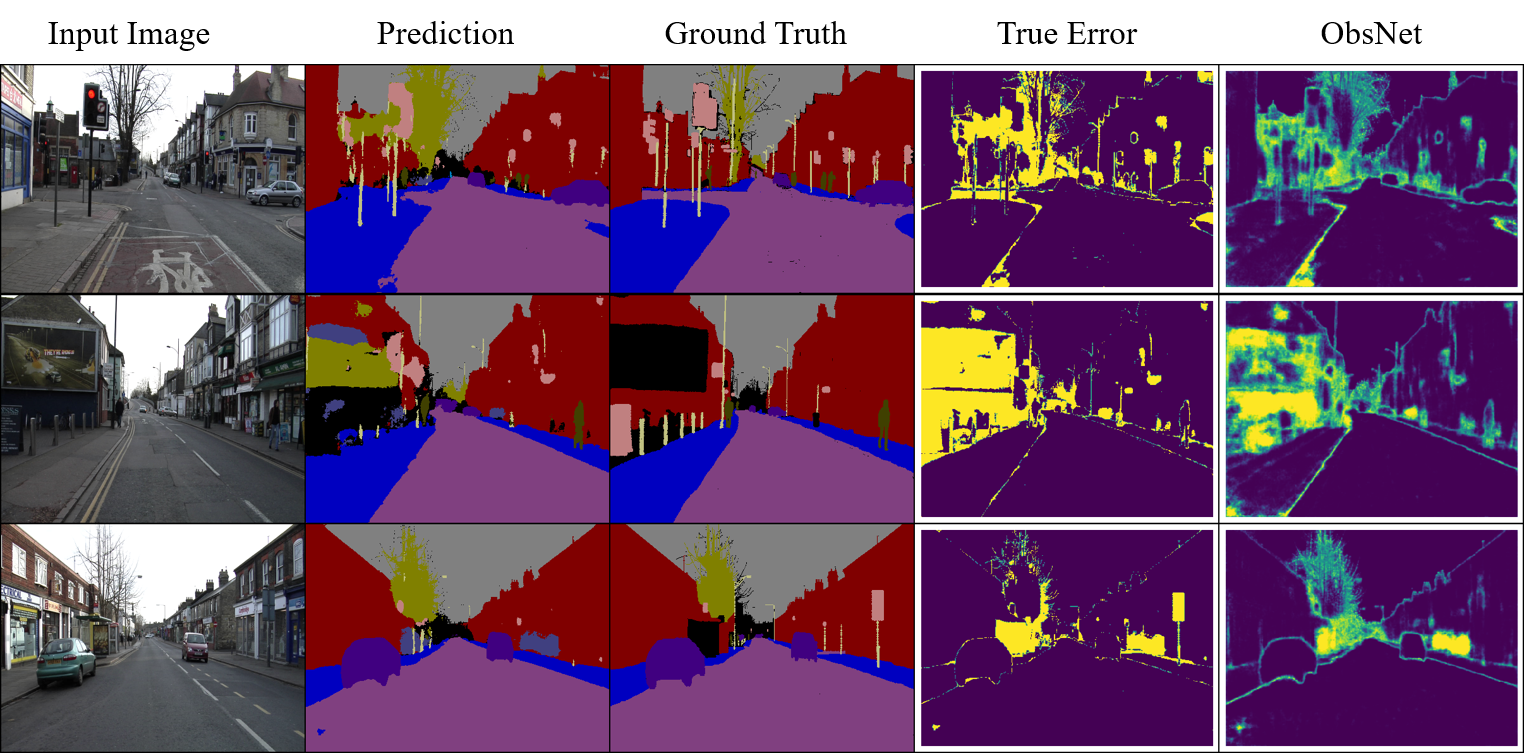}
    \caption{Evaluation of the error detection on the test set of CamVid. \obsnet prediction is close to real errors even without OOD objects.}
    \label{fig:error_detection}
    \vspace{-0.5em}
\end{figure*}

\section{Adversarial Attacks Detector} 
In safety-critical applications like autonomous driving, we know that the perception system has to be robust to adversarial attacks. Nevertheless, training a robust network is costly and robustness come\ab{s} with a \ab{certain} trade-off \ab{to make between} accuracy 
\ab{and run time}. Moreover, the task to \emph{only} detect the adversarial attack could be sufficient as we can rely on other sensors (LiDAR, Radar, etc.). 
\ab{Although, in this work we do not focus on Adversarial Robustness, empirically we note that} \obsnet can detect an attack. 
\ab{To some extent this is expected as} we explicitly train the observer to detect adversarial attacks, thanks to the $\LAA$. 

Indeed, our observer can detect the area where the attack is performed, whereas the \mcdropout is overconfident. Furthermore, in \autoref{tab_eval_adv}, we evaluate the adversarial attack detection of several methods. We apply a FGSM attack in a local square patch on each testing image. Once again, we can see that our observer is the best method to capture the perturbed area.

\end{document}